\documentclass[acmtog,nonacm]{acmart}
\acmSubmissionID{216}
\usepackage{booktabs} % For formal tables
\usepackage{multirow}
\usepackage{tabularx}
\usepackage{wrapfig}
\usepackage{mathtools}
\usepackage{lipsum}
\usepackage{footmisc}

\usepackage[ruled]{algorithm2e} % For algorithms

\SetAlFnt{\small}
\SetAlCapFnt{\small}
\SetAlCapNameFnt{\small}
\SetAlCapHSkip{0pt}

\begin{document}
\def\negativevspace{}	
\newcommand{\TODO}[1]{{\color{red}{[TODO: #1]}}}
\newcommand{\rh}[1]{{\color{green}#1}}
\newcommand{\rz}[1]{{\color{magenta}#1}}
\newcommand{\lzz}[1]{{\color{blue}#1}}
\newcommand{\new}[1]{{#1}}
\newcommand{\phil}[1]{{\color[rgb]{0.2,0.8,0.2}#1}}
\newcommand{\jy}[1]{{\color[rgb]{0.7,0.2,0.0}#1}}
\newcommand{\edward}[1]{{\color[rgb]{0.7,0.2,0.7}#1}}
\newcommand{\ednote}[1]{{\color[rgb]{0.7,0.2,0.7}ED: #1}}
\newcommand{\para}[1]{\vspace{.05in}\noindent\textbf{#1}}
\def\ie{\emph{i.e.}}
\def\eg{\emph{e.g.}}
\def\etal{{\em et al.}}
\def\etc{{\em etc. }}
\newcolumntype{C}[1]{>{\centering\arraybackslash}p{#1}}

 \newcommand{\ourmethod}{CLIPXPlore}

\title{CLIPXPlore: Coupled CLIP and Shape Spaces for 3D Shape Exploration}

\author{Jingyu Hu}
\authornote{Both authors contributed equally to the paper.}
\author{Ka-Hei Hui}
\authornotemark[1]
\author{Zhengzhe Liu}
\affiliation{%
	\institution{The Chinese University of Hong Kong}
        \country{HK SAR, China}}
\author{Hao Zhang}
\affiliation{%
	\institution{Simon Fresor University}
        \country{Canada}}
\author{Chi-Wing Fu}
\affiliation{%
	\institution{The Chinese University of Hong Kong}
        \country{HK SAR, China}}
\renewcommand\shortauthors{Hu et al.}

\begin{abstract}
This paper presents CLIPXPlore, a new framework that leverages 
a vision-language model to guide the exploration of the 3D shape space.
Many recent methods have been developed to encode 3D shapes into a learned latent shape space to enable generative design and modeling.
Yet, existing methods lack effective exploration mechanisms,
despite the rich information.
To this end, we propose to leverage CLIP, a powerful pre-trained vision-language model,
to aid the shape-space exploration.
Our idea is threefold.
First, we couple the CLIP and shape spaces by generating paired CLIP and shape codes through sketch images and training a mapper network to connect the two spaces.
Second, to explore the space around a given shape,
we formulate a co-optimization strategy to search for the CLIP code that 
better matches the geometry of the shape.
Third, we design three exploration modes, binary-attribute-guided, text-guided, and sketch-guided, to locate suitable exploration trajectories in shape space and induce meaningful changes to the shape.
We perform 
a series of experiments to quantitatively and visually compare~\ourmethod~with 
different baselines in each of the three exploration modes, showing that~\ourmethod~can produce many 
meaningful exploration results that cannot be achieved by the 
existing solutions.

\end{abstract}

\maketitle

\section{Introduction}
\label{sec:intro}

Generative design and modeling of 3D shapes has been a long-standing problem in computer graphics. Since the design goals
are often open-ended, even vague, the modeling process is naturally {\em iterative\/} and {\em exploratory\/}. Early
works have already studied exploratory modeling as a means to empower novice users to create high-quality 3D 
shapes~\cite{Talton:TOG:2009,Chaudhuri:TOG:2010}. Even earlier, both language~\cite{Coyne:SIGGRAPH:2001} and sketching~\cite{Igarashi:SIGGRAPH:1999}
interfaces have been developed to best connect the user and the 3D modeling endeavor. 

With the proliferation of large-scale 3D data,
the modern approach is to learn a high-dimensional 3D shape latent space, where each latent code can be decoded into a 3D shape, 
and perform shape generation to explore the latent space.

% \begin{teaserfigure}
\begin{figure}[t]
\vspace*{-1mm}
  \centerline{\includegraphics[width=0.95\linewidth]{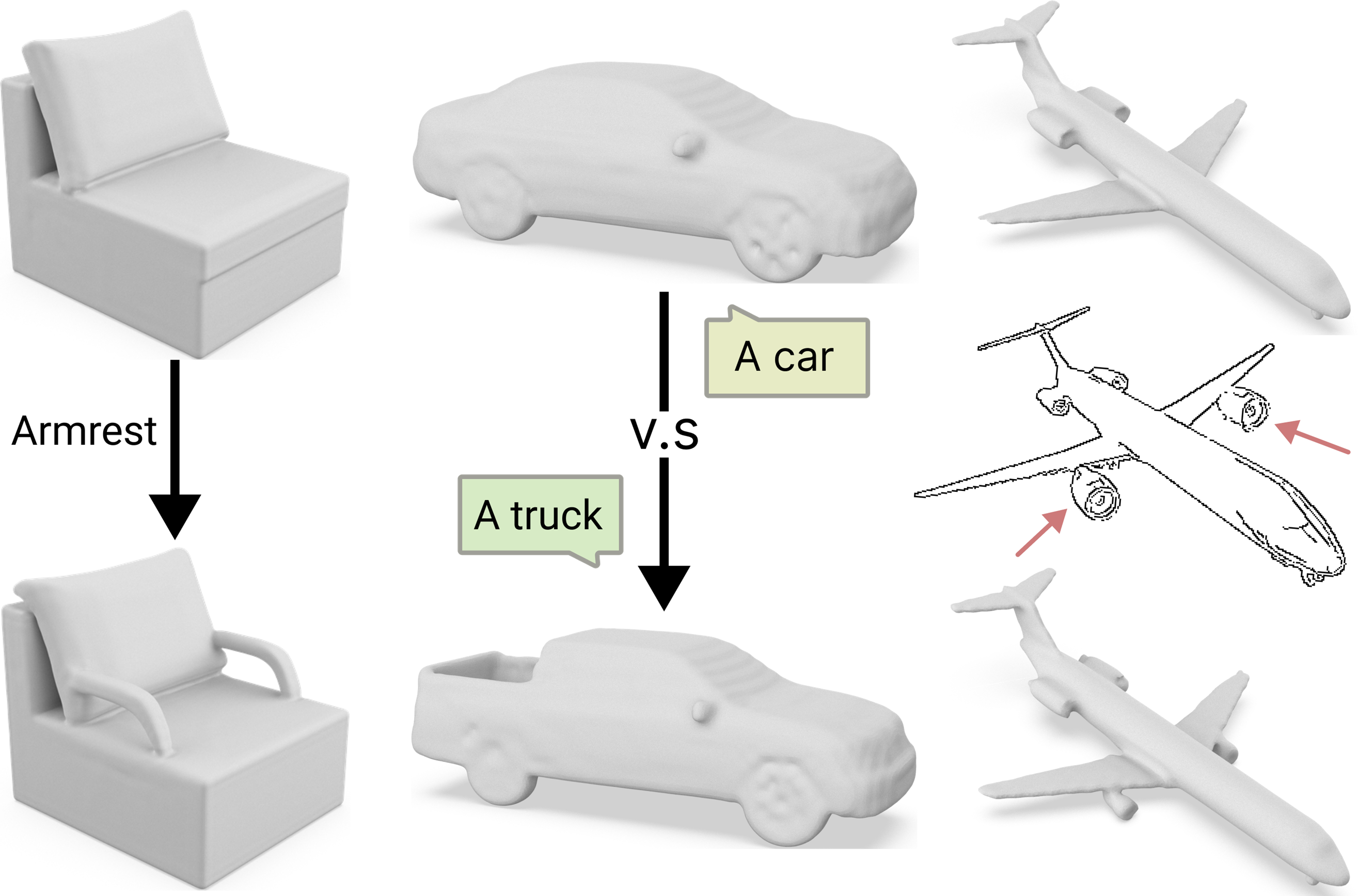}}
\vspace*{-3.5mm}
\caption{\ourmethod~demonstrates new capabilities of leveraging a vision-language model to explore the 3D shape space: binary-attribute-guided (left), text-guided (middle), and sketch-guided (right).
Top row shows input shapes, whereas bottom row shows associated shapes from exploration.
}
\vspace*{-3.5mm}
\label{fig:teaser}
\end{figure}
% \end{teaserfigure}

Many recent works have focused on improving the quality of the constructed shape latent spaces, e.g., in term of the plausibility
and diversity of the decoded shapes~\cite{hertz2022spaghetti,zheng2022sdfstylegan,mittal2022autosdf,zhang20223dilg}. 
However, the latent space exploration has so far been restricted to linear interpolation~\cite{chen2019learning,park2019deepsdf,hertz2022spaghetti} between a given 
shape pair
or tracing along random directions~\cite{gao2022get3d} or a direction that leads to the most significant changes in the 
source shape~\cite{chiu2020human}.
Aside from a lack of fine-grained semantic control, whether for locating a shape code with desirable properties or 
for identifying 
a search direction to impose
meaningful shape changes, these modes of exploration do not 
naturally connect
to the user as a language or 
sketching interface would.

Lately, the research community has introduced several large-scale 
pre-trained vision-language
models, most notably CLIP~\cite{radford2021learning},
that can encode texts and images into a common latent space. 
These models have demonstrated much success in solving various challenging zero-shot 2D and 3D tasks, such as image generation~\cite{ramesh2022hierarchical,rombach2022high,saharia2022photorealistic}, shape generation~\cite{jain2022zero,sanghi2022clip}, and 
shape understanding~\cite{zhang2022pointclip,zhu2022pointclip}. 

Motivated by these advances, we aim to leverage the learned data priors from such encoders, specifically the CLIP model, and to 
explore the pre-trained shape space for a given input 3D shape along a local trajectory to obtain a new one with desired properties.
We expect a well-designed CLIP-enabled latent space to not only offer the potential for 
high-level
semantic control but also bridge the 3D shape 
generative process with natural language and image-space manipulation, without relying on an expensive paired text-shape data.
To this end, 
we need to achieve the following objectives while addressing the ensuing challenges:
\begin{itemize}
    \item Foremost, connect the CLIP space for {\em image+text\/} and a pre-trained latent space for {\em 3D shapes\/}. With the exploration happening in the former and shape generation/decoding obtained by the latter, we must intricately couple each CLIP latent code with a corresponding shape code, so that tracing a linear direction in the CLIP space is equivalent to traversing a trajectory in the shape latent space.
    \item Identify a starting code in the CLIP space, for a given 
    shape to be explored. This CLIP code is paired with a shape code and must faithfully represent the input shape.
    \item Determine an exploration direction in the CLIP latent space, given a user-specified exploration {\em condition\/}, 
    e.g., a language command.
    By tracing along this direction from the starting CLIP code, we would obtain a corresponding shape-space trajectory that satisfies the condition.
\end{itemize}

To achieve 
the objectives, we develop a novel training and inference framework that presents three 
associated technical contributions.
First, we pair each 3D shape in the training set with a {\em rendered sketch image\/} and train a {\em mapper network\/} called CLIP2Shaper, which maps each encoded CLIP code to its associated shape code, ensuring a coherent connection between the CLIP and shape latent spaces.
Second, we {\em co-optimize\/} the CLIP code naively encoded from the sketch 
and its 
associated shape code to more faithfully represent the input
shape for exploration.
Lastly, we design three 
exploration modes with varying multi-modal conditions to identify exploration directions in CLIP space.
The reasons for choosing sketch images to form the CLIP space are two folds. First, the rich geometry and 
structure information conveyed by rendered sketches can well capture most of the relevant shape characteristics for our exploration task. 
Second, the resulting CLIP space allows both language and user sketching as a means to specify exploration conditions.

From the many presented results,~\eg, Figure~\ref{fig:teaser}, we demonstrate the capability of our proposed framework, coined~\ourmethod, which enables exploration of the 3D shape space to produce many meaningful variations of an input shape based 
on binary attributes, text commands, and 2D sketching.
We compared our framework quantitatively and qualitatively with various baselines in each exploration mode and showed its superior performance.
Additionally, a user evaluation showed a preference for our explored results over those produced by existing methods.
Lastly, we conducted various model analyses to confirm the coherent connection of the latent spaces and the effectiveness of the co-optimization strategy.

%\vspace*{-1mm}
\section{Related Work}
\label{sec:rw}

Exploratory design for 3D shape modeling has been studied in computer graphics long before the deep learning era~\cite{Talton:TOG:2009,Chaudhuri:TOG:2010,Yang:TOG:2011,Ovsjanikov:TOG:2011,Xu:TOG:2012,Umetani:TOG:2012,Averkiou:CGF:2014,Zheng:TVCG:2016} and the same can be said about
language-driven~\cite{Coyne:SIGGRAPH:2001,Ma:TOG:2018} and sketch-based~\cite{Igarashi:SIGGRAPH:1999,Olsen:2009,Ding:2016} modeling. In this section, we mainly focus on modern learning-based
approaches that are most relevant to CLIPXplore, especially those involving image-space and shape latent-space exploration and multi-modal shape modeling.

\vspace*{-3pt}
\paragraph{Image latent-space exploration}
Based on the advanced image generation frameworks, such as StyleGAN~\cite{karras2019style}, existing works~\cite{abdal2019image2stylegan, karras2021alias, karras2020analyzing, tov2021designing} construct a latent feature space for 
reconstructing unseen images.
Leveraging a trained latent space, they further enable exploring the latent space by finding an interpretable direction. 
Existing approaches 
can roughly be classified into supervised, unsupervised, and text-guided. 
Supervised methods~\cite{abdal2021styleflow, goetschalckx2019ganalyze, shen2020interpreting} require annotating a set of generated samples. 
By training a classifier using the annotated samples,
they derive an interpretable direction to guide the image modification.
Yet, these methods require tedious data annotations and have limited flexibility in exploring the space.
Unsupervised methods~\cite{shen2021closed, voynov2020unsupervised, cherepkov2021navigating, harkonen2020ganspace, wei2021orthogonal} find the interpretable direction without using annotated samples,~\eg, by a PCA decomposition on the network weights or on the latent codes.
Still, both approaches struggle to identify the interpretable direction that leads to the desired changes.
To improve the controllability, various works~\cite{xia2021tedigan, patashnik2021styleclip, gal2022stylegan, abdal2022clip2stylegan} adopt text as a condition to modify images.
Very recently,~\cite{patashnik2021styleclip, abdal2022clip2stylegan} leverage large-scale pre-trained vision-language models,~\eg, CLIP~\cite{radford2021learning}, for text-guided image-space exploration without requiring paired text-image data.

Despite the success of 2D methods,
it remains unclear how one may explore the 3D latent space.
Recently,~\cite{zheng2022sdfstylegan} directly extend supervised methods to handle 3D latent space.
Yet, they require extensive annotations, thus limiting the practical usage.
We also observe degraded performance when directly exploring the 3D latent space.
To fill this gap, we present a general
framework that takes advantage of a pre-trained vision-language model to help identify the exploration trajectory
in the latent shape space.

\begin{figure*}[t]
	\centering
	\includegraphics[width=0.99\linewidth]{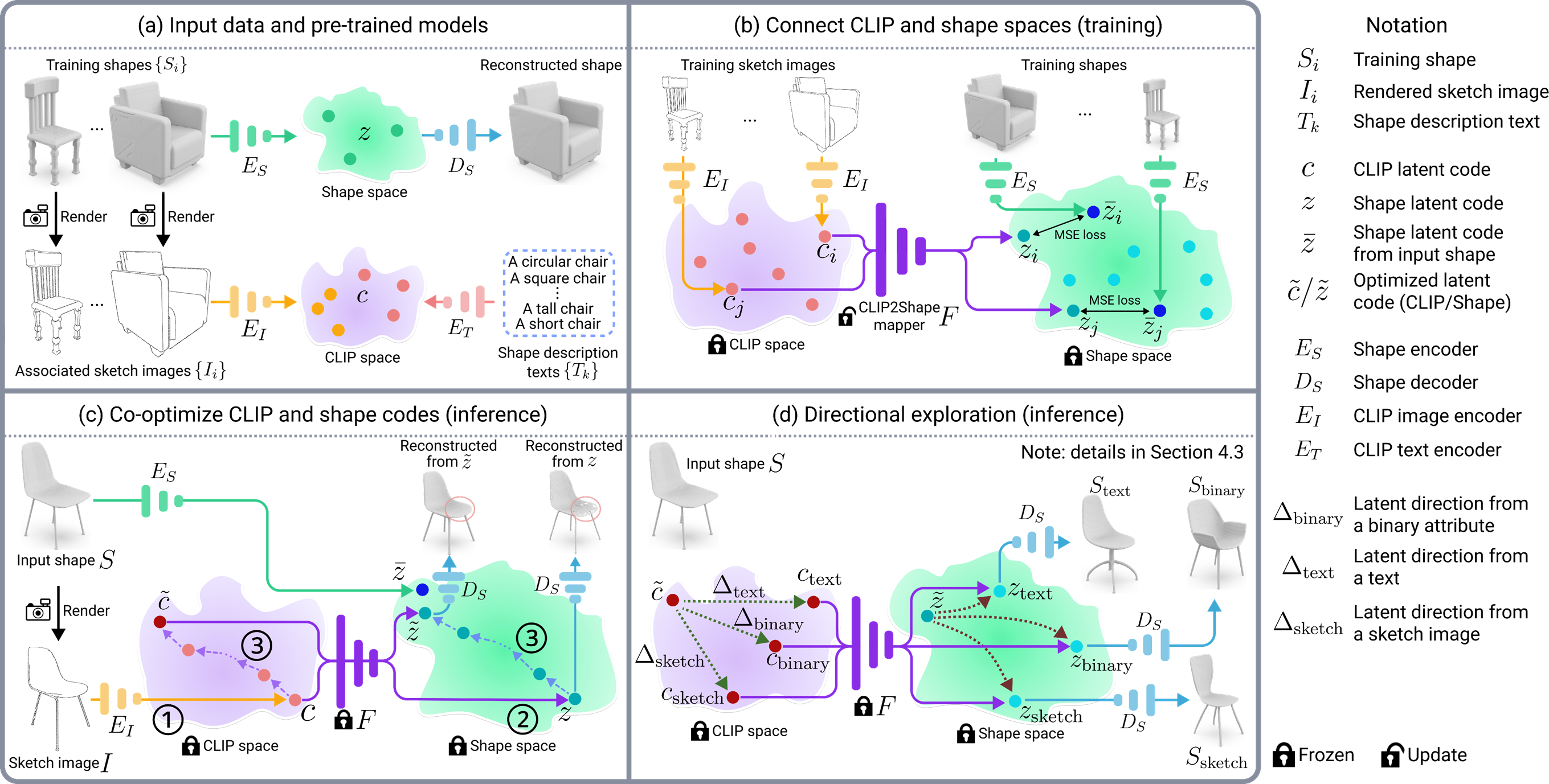}
	\vspace*{-0.75mm}
	\caption{
        Overview of~\ourmethod.
        (a) summarizes our employed data,~\ie, training shapes $\{S_i\}$, rendered sketch images $\{I_i\}$, and shape description texts $\{T_k\}$, as well as pre-trained models, $E_S$, $D_S$, $E_I$, and $E_T$.
        (b) connects the CLIP and shape spaces by preparing paired CLIP codes $\{c_i\}$ and shape codes $\{\bar{z}_i\}$ via the rendered sketch images and then by training the CLIP2Shape mapper $F$: 
        $c_i \rightarrow z_i$, where $z_i$ should be close to $\bar{z}_i$.
        (c) co-optimizes the CLIP code $c$ and its coupled shape code $z$ to find a better CLIP code $\tilde{c}$ whose mapped shape code $\tilde{z}$ is closer to $\bar{z}$.
        (d) locates the latent direction in the CLIP space, $\Delta_{\text{text}}$, $\Delta_{\text{binary}}$, or $\Delta_{\text{sketch}}$, associated with the given condition, allowing us to trace along the direction to modify the codes ($\tilde{c}$ and $\tilde{z}$) and explore the shape space.
        }
	\label{fig:overview}
	\vspace*{-0.35mm}
\end{figure*}

\vspace*{-3pt}
\paragraph{Shape latent-space exploration}
Recently, numerous methods have been designed to learn a latent space for reconstructing and generating shapes
using various 3D representations,~\eg, 
voxels~\cite{wu2016learning, smith2017improved}, point clouds~\cite{achlioptas2018learning, gal2020mrgan, hui2020progressive, xie2021generative}, meshes~\cite{wang2018pixel2mesh, groueix2018, chandraker2020neural, chen2020bsp, gao2022get3d}, and implicit functions~\cite{chen2019learning, park2019deepsdf}.
Encoding 3D shapes of irregular structures and complex topologies into a compact latent space can help improve the controllability when generating 3D shapes.
For example, by leveraging the latent space,~\cite{park2019deepsdf,chen2019learning,hertz2022spaghetti} semantically mix different shapes by interpolating their latent codes.
Others produce variations of a shape by permuting its latent code~\cite{gao2022get3d} towards a random direction or by exploring in the direction that leads to most changes~\cite{chiu2020human}.

However, existing approaches have very limited controllability.
For example, it is challenging to obtain the interpretable direction in their latent space 
that modifies
the shape for a given user-specified condition.
To this end,
we propose the first framework by leveraging the pre-trained CLIP space to enhance the controllability when exploring the 3D shape space, such that 
we can find an exploration trajectory that satisfies the desired changes.

\vspace*{-3pt}
\paragraph{Multimodal 3D shape modeling}
To improve the controllability, many methods attempt to incorporate additional-modality inputs
such as texts and sketches into the 3D shape modeling process.
Particularly, several researchers adopted paired text-shape data to generate shapes that match the given text descriptions~\cite{chen2019text2shape,liu2022towards,fu2022shapecrafter}.
Recently, some methods further leverage
language descriptions
about the difference between a shape pair to give hints for text-guided shape manipulation~\cite{achlioptas2022changeIt3D,huang2022ladis}. 
Yet, preparing paired text-shape data is extremely tedious and expensive.
Like our framework, CLIP-forge~\cite{sanghi2022clip} also uses the CLIP model to generate 3D shapes based on text descriptions without paired data, but it does not aim for shape manipulation under a given condition.

Another line of work explores text-guided 3D shape manipulation by optimizing the vertices of a mesh~\cite{mohammad2022clip,oscartext2mesh,Gao_2023_SIGGRAPH}, hoping to enhance the consistency bewtween its multi-view differentiable renderings and the text description measured by CLIP~\cite{radford2021learning}.
Yet, these approaches often require significant optimization time and the topologies of manipulated shapes are highly restricted by the given mesh.

Different from paired text-shape data,
paired sketch-shape data can be easily obtained by rendering 3D shapes nonphotorealistically. 
Also, sketch images are relatively easy to edit, making it an effective medium for controlling the generated shapes. 
To this end, several works~\cite{zhang2021sketch2model, guillard2021sketch2mesh, gao2022sketchsampler, cheng2022cross, delanoy20183d, li2018robust, han2020reconstructing, zhong2020towards} leverage sketch images to generate and edit 3D shapes.
However, their generative fidelity and controlling accuracy are far from satisfactory, thereby limiting their practical usage.
A concurrent work~\cite{zheng2023lasdiffusion} exploits the diffusion model to enhance the visual quality of results, yet it focuses only on the generation task, given a single-modality condition (sketch image).

Very recently, some works~\cite{mittal2022autosdf, cheng2022sdfusion, li20233dqd} start to collectively exploit multiple modalities, such as text and image, taking them as conditions
to improve the controllability in the generation pipeline. 
However, these methods are still limited by the scale of the paired text-image-shape data,
which is typically very costly to prepare.
In this work, we present a general CLIP-induced framework,
in which we can take text prompts or sketch images to modify 3D shapes.
Our approach can find relevant exploration 
trajectories in the latent shape space according to the given condition, 
without requiring paired text-shape data.

% \input{tables/notation}

%\vspace{-1mm}
\section{Overview}
\label{sec:overview}

Given an input shape and 
a condition (text, binary attribute, or sketch image) to modify it, we aim to identify an exploration trajectory in the shape space that can induce relevant changes to the shape.
Figure~\ref{fig:overview} provides an overview of our~\ourmethod~framework.

Figure~\ref{fig:overview}(a) summarizes the employed input data and pre-trained models.
The input data includes a set of training shapes $\{ S_i \}$ and a set of shape description texts $\{ T_k \}$.
We use $\{ T_k \}$ only in inference and
render each training shape $S_i$ into a sketch image $I_i$; see Section~\ref{ssec:train} for details.
Besides the input data, we employ four pre-trained models: $E_S$/$D_S$ for encoding/decoding the shapes, $E_I$ for encoding the sketch images, and $E_T$ for encoding the shape description texts.

With the introduced inputs and pre-trained spaces, our proposed \ourmethod~framework consists of three major components:

(i) <training> {\em Connect CLIP and shape spaces.}
To connect the two pre-trained spaces, %on the one hand, 
we encode each rendered sketch image $I_i$ into the CLIP space to obtain CLIP code $c_i$ for the corresponding training shape $S_i$. 
At the same time, we directly encode each training shape $S_i$ to the shape space to obtain shape code $\bar{z}_i$.
With the associated code pairs, $\{c_i\}$ and $\{\bar{z}_i\}$, we can then train the CLIP2Shape mapper $F$ to connect the two spaces; see Figure~\ref{fig:overview}(b).
In short, we freeze the two latent spaces and update the mapper by encouraging the mapped code $z_i$ to be close to shape code $\bar{z}_i$ for each training shape,~\ie, $z_i = F(c_i) \approx \bar{z}_i$.
Details will be provided in Section~\ref{ssec:train}.

(ii) <inference> {\em Co-optimize CLIP and shape codes.}
Given a shape for exploration, we first encode its sketch image to obtain an initial CLIP code $c$; see step (1) in Figure~\ref{fig:overview}(c). 
Then, we use the trained mapper to obtain a paired shape code $z$ in step (2). 
However, the code $z$ may not faithfully reconstruct the original shape due to information loss from the rendered sketches; compare the given shape $S$ and the shape reconstructed from $z$. 
Hence, we co-optimize 
the CLIP code $c$ and shape code $z$, in step (3),
to find a better CLIP code $\tilde{c}$ whose mapped shape code $\tilde{z}$ is closer to the shape code $\bar{z}$ 
by directly encoding the input shape; see Section~\ref{ssec:start_point_opt}.

(iii) <inference> {\em Directional exploration.}
After obtaining the optimized CLIP code $\tilde{c}$, we need to locate the latent direction along which to modify it subject to a given condition.
To this end, our unified framework 
supports multiple 
conditions, including a text, binary attribute, or sketch, as shown in Figure~\ref{fig:overview}(d).
We first locate a latent direction in the CLIP space,~\ie, $\Delta_{\text{text}}$, $\Delta_{\text{binary}}$, or $\Delta_{\text{sketch}}$, that matches the given condition. Then,
we trace along the direction in the CLIP space, map the modified CLIP code to shape space, and produce the modified shape;
see Section~\ref{ssec:dir_explore} for details.

%\vspace{-1mm}
\section{Method}
\label{sec:architecture}

\subsection{Connect CLIP and shape spaces}
\label{ssec:train}

To connect the two spaces,
our key idea is to prepare pairs of associated CLIP code and shape code through the rendered sketch images and then train the CLIP2Shape mapper.
In detail, we employ the Open3D renderer~\cite{zhouopen3d2018} to render each training shape $S_i$, apply canny-edge detector to produce the associated sketch image $I_i$, 
then encode the sketch image into CLIP code $c_i = E_I(I_i)$.
On the other hand, we directly encode each training shape $S_i$ into shape code $\bar{z}_i = E_{S}(S_i)$.
Note also that by adopting a sketch image as input using the pre-trained CLIP image encoder $E_I$, we can leverage the sketch as a condition for exploring the latent shape space.

With the paired codes, $c_i$ and $\bar{z}_i$, 
the task of the CLIP2Shape mapper network is to regress shape code $z \in R^n$ given the input CLIP code $c \in R^m$, where $m$ and $n$ are the dimensions of the CLIP and shape spaces, respectively.
We build the mapper by adopting a simple yet effective network architecture.
The architecture consists of eight MLP layers with leaky-ReLU activation, and between adjacent layers, we employ the skip-connection to enhance the network performance, as inspired by~\cite{he2016deep}.
To train the CLIP2Shape mapper, we utilize an $L_2$ regression loss as follows:
\begin{equation}
    L = || z_i - \bar{z_i}||^2
    \label{eq:c2s_training}
\end{equation}
where $z_i = F(E_{I}(I_i))$ and $\bar{z}_i =E_{S}(S_i)$. 
In the training, we update only the CLIP2Shape parameters while fixing encoders $E_I$ and $E_S$.
Also, as the architecture is lightweight, the training is highly efficient, completing in around one hour on a single GPU.

\begin{figure}[!t]
	\centering
	\includegraphics[width=0.99\columnwidth]{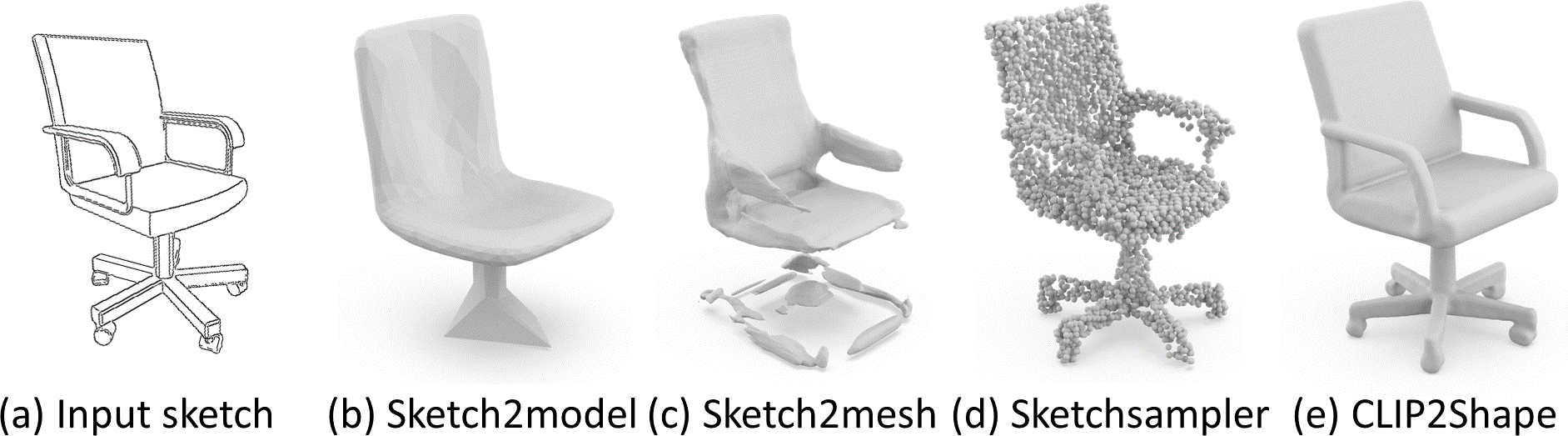}
        \vspace{-3mm}
	\caption{
     Using the trained CLIP2Shape mapper, we can obtain shape codes that lead to good 3D reconstructions (e) with nontrivial structures,~\eg, vertical bars, which are challenging for existing methods (b-d),
     showing that the CLIP2Shape mapper can well connect the CLIP and shape spaces.
}
	\label{fig:recon_comp}
    \vspace{-3mm}
\end{figure}

Essentially, the CLIP2Shape mapper is like a 3D reconstruction engine, producing shape codes (from single-view sketch images) that can be decoded into 3D shapes.
So, we can evaluate how well it connects the two spaces by studying the 3D reconstruction quality from the mapped shape codes.
Figure~\ref{fig:recon_comp} shows 3D reconstructions produced from an example sketch image.
Our approach is able to reconstruct more complex structures just from the mapped code.
This is challenging even for some existing methods.
From these results, we can see that our trained CLIP2Shape mapper can establish a good connection between the CLIP and shape spaces, serving as a good foundation for exploring the shape space later on.
More detailed quantitative evaluation 
is provided in supplementary materials.

\begin{figure}[t]
	\centering
	\includegraphics[width=0.9\linewidth]{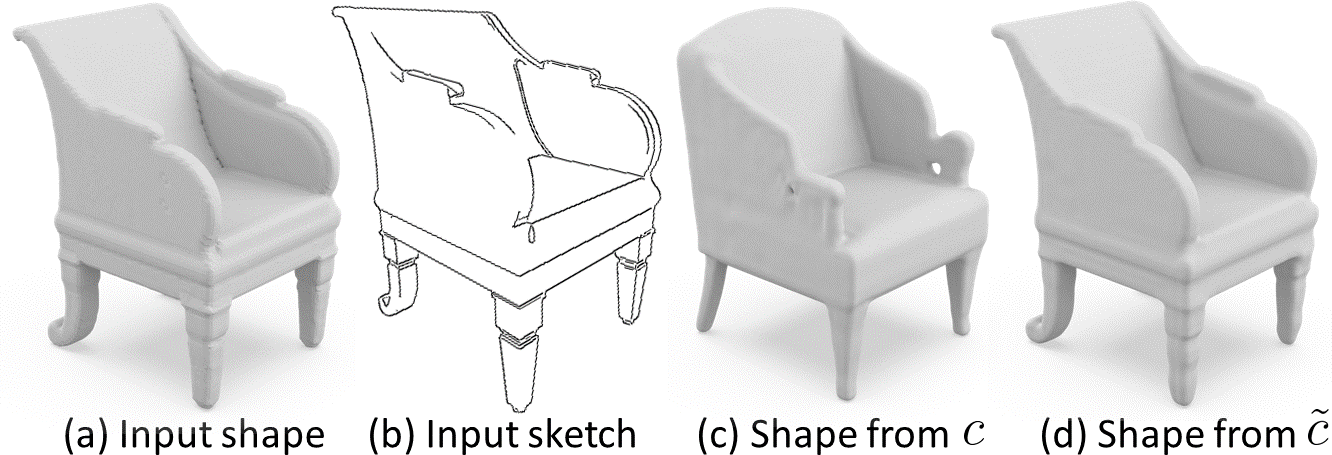}
        \vspace{-3mm}
	\caption{Using the optimized CLIP code $\tilde{c}$, we can reconstruct the shape in (d), which is more faithful to the input (a) with more fine details, compared with the shape reconstructed from CLIP code $c$ without our co-optimization (c).
  Please refer to the supplementary material for more visual ablations.
 }
	\label{fig:start_op_ab}
 \vspace{-3mm}
\end{figure}

\begin{figure*}[t]
    \begin{minipage}[t]{.33\textwidth}
        \centering
        \includegraphics[width=0.99\textwidth]{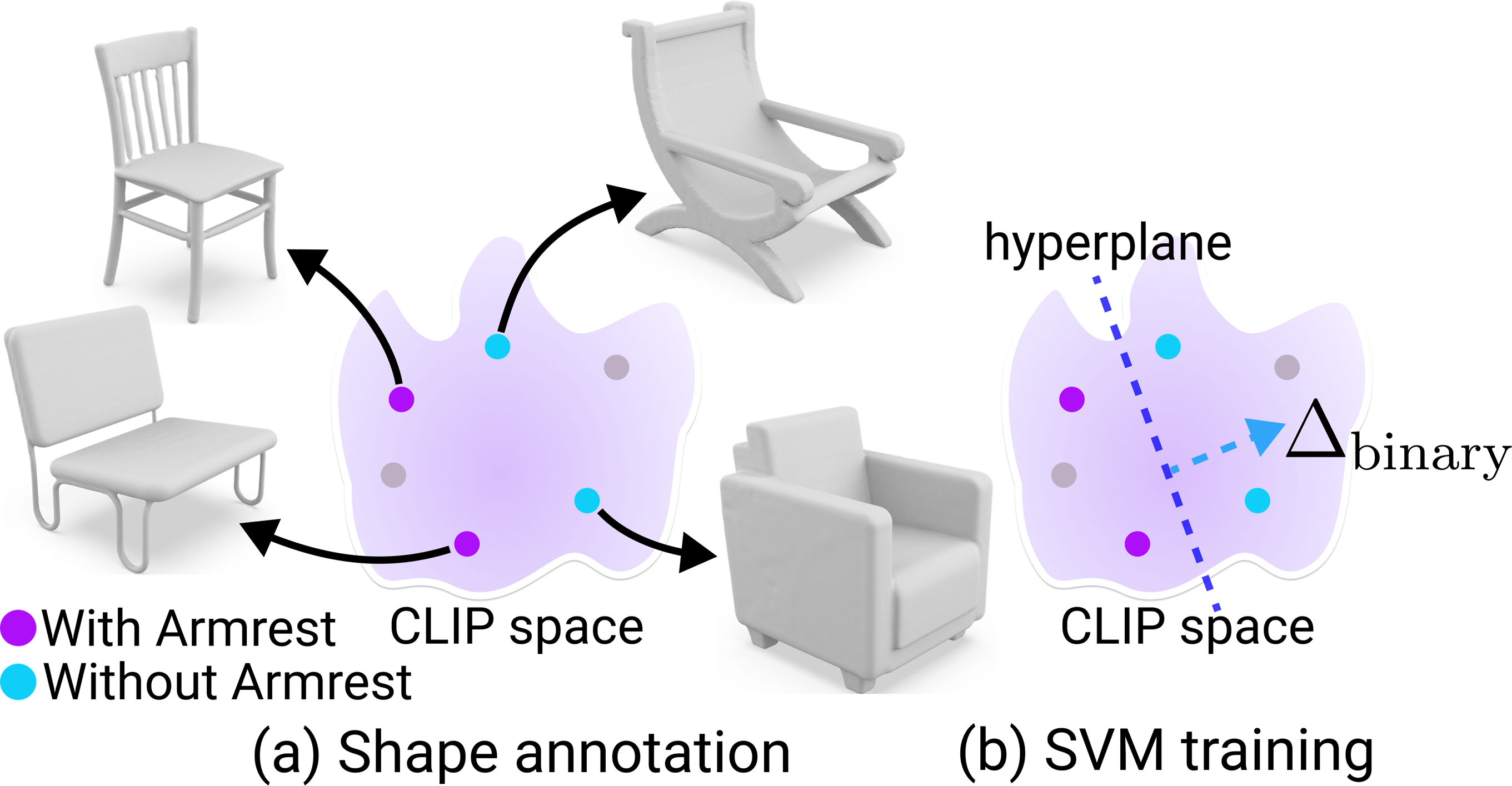}
        \vspace*{-6mm}
        \caption{Binary-attribute-guided exploration.
                (a) We first annotate shape samples with/without 
                a certain attribute,~\eg, armrest.
                (b) We then train a linear SVM to obtain the latent exploration direction.}
    \label{fig:binary_overview}
    \end{minipage}
    \hfill
    \begin{minipage}[t]{.29\textwidth}
        \centering
        \includegraphics[width=0.99\linewidth]{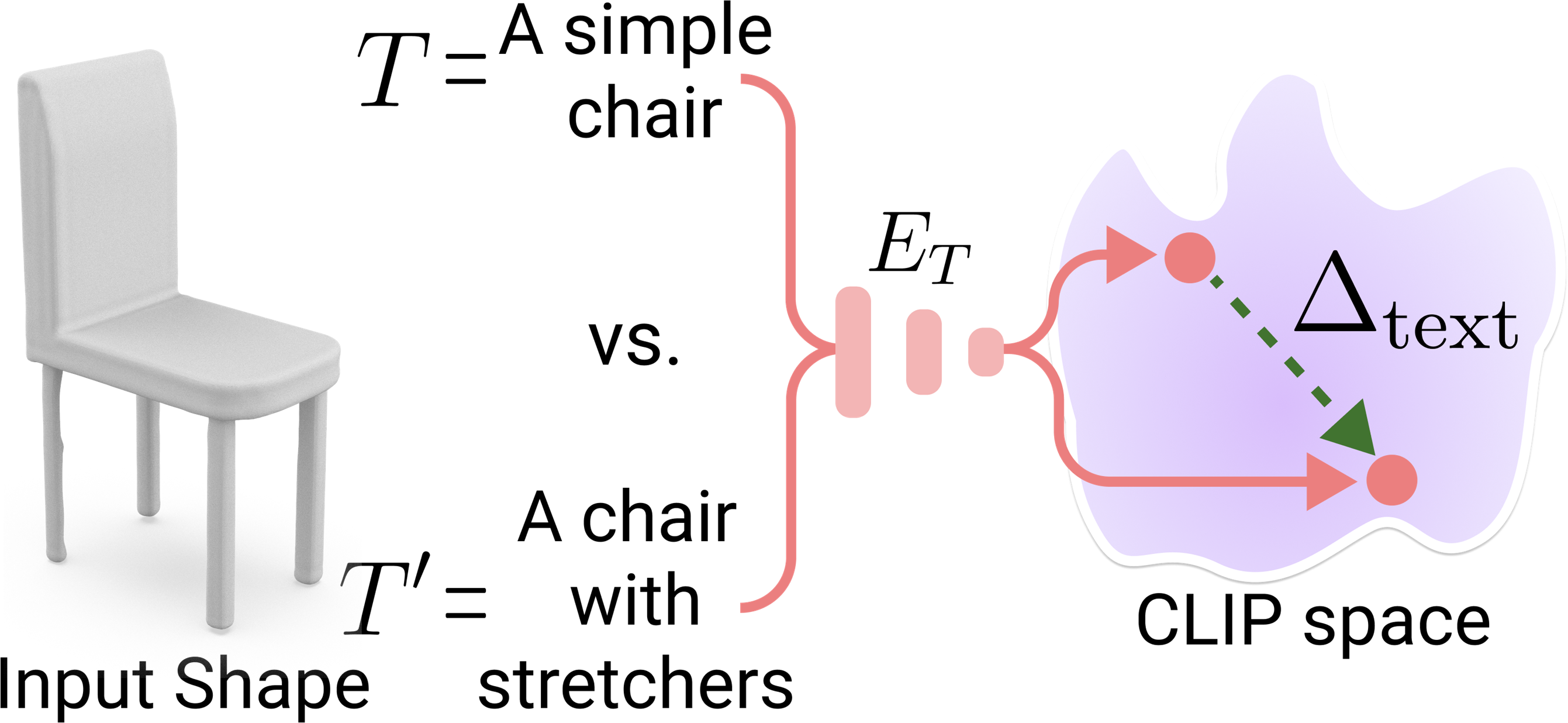}
        \vspace*{-6mm}
        \caption{
        Text-guided exploration.
        Given a shape description pair, such as $T$ and $T'$ shown above, we locate the latent direction as the vector between their CLIP codes.
        }
        \label{fig:text_overview}
        \end{minipage}
        \hfill
    \begin{minipage}[t]{.35\textwidth}
        \centering
        \includegraphics[width=0.99\textwidth]{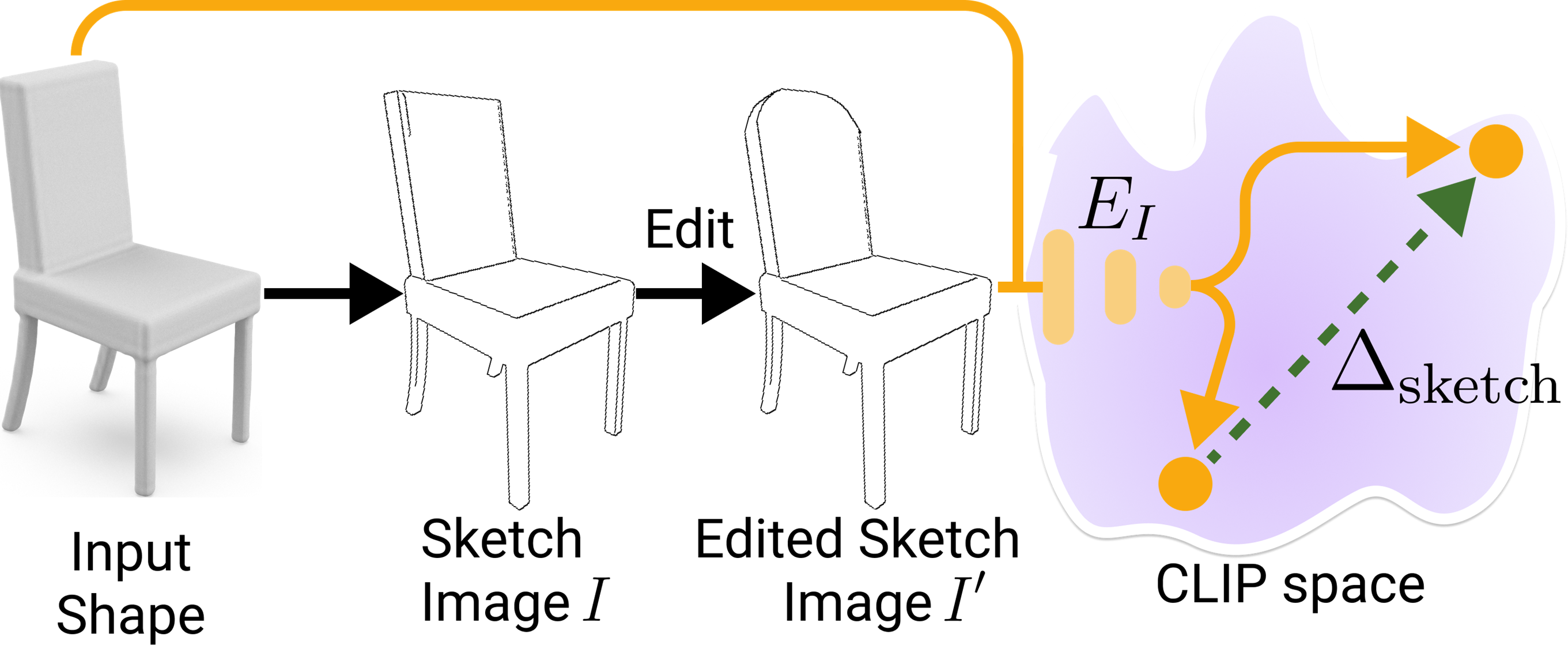}
        % \vspace{-1mm}
        \vspace*{-6mm}
        \caption{
        Sketch-guided exploration.
        The input sketch image and edited sketch image together form the condition, from which we encode them into the CLIP space and locate the latent exploration direction.
        }
        \label{fig:sketch_overview}
    \end{minipage}  
    \vspace{-2mm}
\end{figure*}

\subsection{Co-optimize CLIP and shape codes}
\label{ssec:start_point_opt}

With the CLIP2Shape mapper,
we are ready to start the shape exploration (inference) on the given shape.
To do so, we first need to find a CLIP code that serves as the starting point in the exploration.

A straightforward solution is to just render the given shape, encode its rendered sketch image into a CLIP code, which is $c$, and take $c$ as the starting point.
Though $z=F(c)$ is a fairly good representation of the input shape, it might still exhibit small variations from the shape code $\bar{z}$, a direct encoding of the given shape.
Figure~\ref{fig:start_op_ab}(a) vs. (c) show a visual comparison between the input shape and the shape reconstructed from $z=F(c)$.
The underlying reason is due to the information loss when rendering the shape and encoding its sketch image.
Hence, to enhance the shape exploration quality, 
we propose to further co-optimize $c$ and its paired shape code $z$ (see Figure~\ref{fig:overview}(c)), such that we can locate $\tilde{c}$ in the CLIP space, with $F(\tilde{c})$ closer to shape code $\bar{z}$.
Using this approach, we can find a better starting point $\tilde{c}$ to aid the shape exploration.

In detail, we 
co-optimize the CLIP code $c$ and shape code $z$ iteratively.
First, we set $c_1 = c$ and $z_1 = z$.
Then, at iteration $t$,  
we modify the codes using the regression loss between shape code $z_t$ and target code $\bar{z}$ by using Equation~(\ref{eq:c2s_training}).
Since the Clip2Shape mapper is fully differentiable, we can back-propagate the gradient vector of this loss to updated CLIP code $c_t$ to $c_{t+1}$ in the CLIP space, such that the mapped code $z_{t+1} = F(c_{t+1})$ is closer to $\bar{z}$.
During this process, the CLIP2Shape mapper is fixed, so the learned connection between the CLIP and shape spaces is not disrupted.
Also, the iteration is fast, without model re-training or fine-tuning.
Hence, we can quickly find the optimized CLIP code $\tilde{c}$ with minimized distance between $F(\tilde{c})$ and $\bar{z}$ in the shape space.
Figure~\ref{fig:start_op_ab} (c) vs. (d) show a visual comparison between 
shapes reconstructed from the initial CLIP code $c$ and optimized CLIP code $\tilde{c}$.
Note the finer and more faithful details in the shape reconstructed from $\tilde{c}$.
Additional visual ablation results are presented in the supplementary materials.

\vspace{-3mm}
\subsection{Directional exploration}
\label{ssec:dir_explore}
Next, we need to find the latent direction in CLIP space that matches the given condition, such that we can make use of this direction to induce relevant changes to the given shape and explore the shape space.
Our~\ourmethod~framework provides three exploration modes.
Below, we present how we compute the latent direction from the specified condition in each of the three modes:

\vspace*{1mm}
(i) {\em Binary-attribute-guided exploration.}
To specify the condition in this exploration mode, as Figure~\ref{fig:binary_overview}(a) illustrates, we need a set of positive shape samples (\eg, chairs with armrests) and a contrasting set of negative shape samples (\eg, chairs without armrests), which are subsets of the training set.
One way to obtain the two sets is by manual labeling.
Besides, we can employ the PartNet data~\cite{mo2019partnet}, which provides a part hierarchy that lists the semantic parts in each shape.
In our implementation, we use the latter approach.
In detail, using PartNet, we can find a subset of shapes with certain target semantic part (\eg, armrests) as the positive samples and find another subset without that semantic part as the negative samples.
To locate the latent direction in CLIP space from the two sample sets, we first employ encoder $E_I$ to obtain the encoded CLIP code of each shape, then train a linear support vector machine (SVM) to classify the labels given the codes.
We then obtain a decision boundary,~\ie, a CLIP-space hyperplane that separates the positive and negative codes; see Figure~\ref{fig:binary_overview}(b).
By then, we can find the hyperplane's normal, which is a unit vector in CLIP space, and take it as the latent direction, denoted as $\Delta_{\text{binary}}$, for shape exploration.
Importantly, using this approach, we need sufficient number of labeled samples (empirically, more than $1,000$ samples in each set).
Also, we should balance the number of positive and negative samples.

(ii) {\em Text-guided exploration.}
To specify a text condition for exploration, we need to prepare a pair of shape description texts: (i) an input shape description $T$ that matches the given shape (e.g., a simple chair), and 
(ii) a target shape description $T'$ with certain properties (e.g., a chair with stretches); see,~\eg, Figure~\ref{fig:text_overview}.

Then, we can employ the text encoder $E_T$ to encode each description text, $T$ and $T'$, into a CLIP code, and
obtain the latent direction as the 
vector from $T$'s CLIP code to $T'$'s CLIP code,~\ie, $\Delta_{\text{text}} = E_{T}(T') - E_{T}(T)$.
For the text pair, the input description text should match the given input shape.
Also, we found that using an input shape description that contrasts the target shape description can improve the exploration quality.
Further, since we are exploring a pre-trained shape space, it would be more effective, if the target shape description matches the properties of some shapes in the shape dataset that creates the shape space.

\vspace*{1mm}
(iii) {\em Sketch-guided exploration.}
To form a condition using sketching, as Figure~\ref{fig:sketch_overview} illustrates, we first render the given shape into a sketch image, denoted as $I$.
We then provide the user with an interface to edit this image and produce a new one, denoted as $I'$.
Typical editing operations include curving the chair's back and removing certain part.
After the editing, we can obtain a pair of sketch images that serve as the condition for exploring the shape space.
Using the CLIP image encoder $E_I$, we can encode each sketch image into a CLIP code.
Following the procedure in exploration mode (ii), we can then compute the latent direction as the vector from $I$'s CLIP code to $I'$'s CLIP code: $\Delta_{\text{sketch}} = E_{I}(I') - E_{I}(I)$.
It is important to note that since we are exploring a pre-trained shape space, we cannot handle arbitrarily-edited sketch images. 
However, we can specify more precise changes on the input shape.

\vspace*{2mm}
After obtaining the latent direction for shape exploration, we can trace from CLIP code $\tilde{c}$ along direction $\Delta$, which can be $\Delta_{\text{binary}}$, $\Delta_{\text{text}}$, or $\Delta_{\text{sketch}}$, to obtain CLIP code $c = \tilde{c} + \alpha * \Delta$, where $\alpha$ is how far we trace.
We can then map CLIP code $c$ to shape space using the CLIP2Shape mapper and employ shape decoder $D_S$ to reconstruct the modified shape; see again the procedure in Figure~\ref{fig:overview}(d).

\vspace{-2mm}
\paragraph{Discussion on shape exploration methods.}
Each exploration method has strengths and weaknesses.
Binary-attribute-guided exploration generally induces very accurate changes but the obtained latent directions might not be stable if there are insufficient annotated samples.
Text-guided exploration is more suitable for inducing shape changes related to more global appearance, since the existing vision-language model CLIP may not provide fine-grained information for inducing local and specific changes.
Lastly, sketch-guided exploration allows for precise specification of conditions, but inaccurate drawing might lead to a degraded exploration direction.
Hence, we found from our experiments that using sketching to remove local structures or make local edits usually leads to better results.

\begin{figure*}[h]
\vspace*{-1mm}
  \centerline{\includegraphics[width=0.75\linewidth]{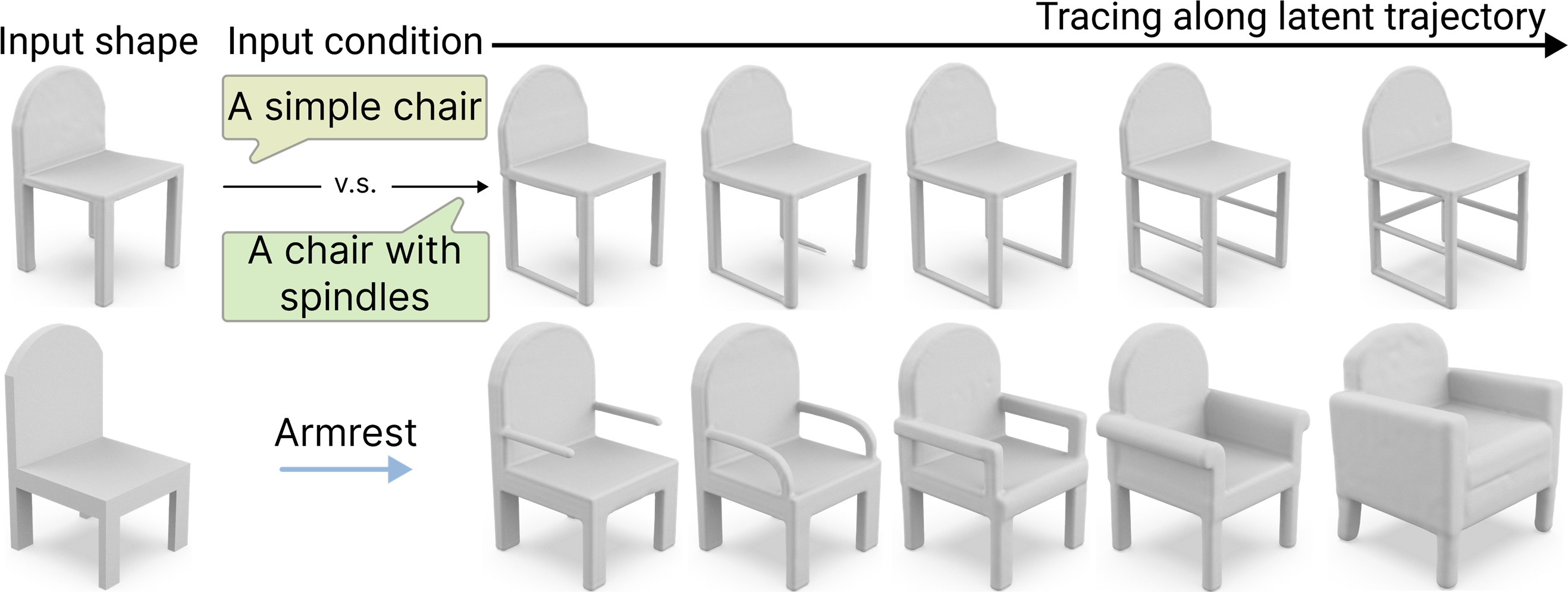}}
\vspace*{-3.5mm}
\caption{
Tracing along the latent direction produced from the text description condition (top row) and from the binary attribute condition (bottom row) allows us to induce shapes that gradually possess the target property (spindles in the top row and armrests in bottom row), as specified by the condition.
}
\label{fig:tracing_demo}
\end{figure*}

\vspace{-2mm}
\paragraph{Discussion on $\alpha$.}
By varying $\alpha$, we can produce a series of shapes along the exploration trajectory in shape space.
Note that instead of exploration direction, we call it exploration trajectory in the shape space, because the latent direction in CLIP space, after mapped to the shape space, is no longer a linear direction.
Figure ~\ref{fig:tracing_demo} shows two visual examples on varying $\alpha$.
In general, a small $\alpha$ leads to a small change on the given shape, whereas a large $\alpha$ produces a shape that could be more different from the input.
Here, a very challenging problem is how to choose a suitable value for $\alpha$.

To do so, we first empirically set range $[\alpha_{min}, \alpha_{max}]$ in a conservative manner.
Then, we propose the following mechanisms:
\begin{itemize}
\vspace*{-1.5mm}
\item[(i)] 
For the case of sketch-guided exploration, our idea is to make use of the edited sketch image and its associated CLIP code, say $c_\text{edited}$.
Procedure-wise, we first obtain multiple shape candidates along $\Delta_{sketch}$ using uniformly-sampled $\alpha$ values, then render these shapes into sketch images and encode these images into the CLIP space.
Hence, we can measure how close each CLIP code is to $c_\text{edited}$ and pick the shape that is the closest to the edited sketch image.
\item[(ii)]
For the cases of binary-attribute- and text-guided exploration, we do not have an image-level prior like the edited sketch image, so we simply set a pre-defined value for $\alpha$.
\item[(iii)]
The above mechanisms are fully automatic, yet may not produce the best result.
So, at the expense of users, we may generate multiple candidate shapes and let the users pick the desired one, following existing works on exploring the latent image space~\cite{shen2020interpreting,shen2021closed}.
Importantly, Our method can complete in the order of minutes, as it does not require fine-tuning and model re-training.
\end{itemize}

\subsection{Implementation details}
\label{sec:implementation}

We employed ShapeNet~\cite{chang2015shapenet} to prepare the training dataset in all our experiments. 
Following the data split in~\cite{chen2019learning}, we use only the training split to supervise the training of CLIP2shape.
For the pre-trained shape encoder and decoder, we adopt the models provided by the authors in~\cite{hui2022neural}, 
which are pre-trained separately per category, following the data split of~\cite{chen2019learning}.
For the CLIP model, we adopt the ViT-B/32 CLIP model~\cite{radford2021learning}, which uses Vision Transformer~\cite{dosovitskiy2020image} as the image encoder and uses the Transformer network~\cite{vaswani2017attention} for the text encoder.

In detail, we train the mapper for 5,000 epochs using the Adam optimizer~\cite{kingma2014adam} with a learning rate of $1e^{-4}$, taking around one hour.
For the co-optimization, we adopt Adam optimizer~\cite{kingma2014adam} to obtain the back-propagation gradient with a learning rate of $2e^{-4}$ for 2,000 iterations, altogether taking only around 20 seconds. 
Lastly, the inference takes around one minute to reconstruct a shape on an RTX 3090 GPU.
{\em We will release code and trained model upon the publication of this work.}

\section{Results and Experiments}

\subsection{Experiment settings}
\label{ssec:exp_setting}
To evaluate the performance of our framework, we compare it with various baselines on each of the three exploration modes:

\paragraph{Binary-attribute-guided.} 
We compare our framework with directly training the SVM in the shape space, as in~\cite{zheng2022sdfstylegan}.
Both the baseline and our method use the same set of annotated samples and follow the same training procedure in Section~\ref{ssec:dir_explore}.
The only difference is that the baseline uses shape codes $\{ z \}$ to train the SVM and directly explore the shape space.
After training the SVMs, we evaluate our method and the baseline on 50 exploration cases (25 chairs and 25 tables), where each case contains an input shape and a binary attribute.
The modified shapes are then produced by tracing along a fixed step size along the exploration direction.

\paragraph{Text-guided.} 
In this exploration mode, we evaluate our framework against three text-based shape manipulation methods: CLIP-Mesh~\cite{mohammad2022clip}, Text2Mesh~\cite{oscartext2mesh}, and TextDeformer~\cite{Gao_2023_SIGGRAPH}.
These methods progressively optimize (deform) an input mesh to better match the target shape description via CLIP~\cite{radford2021learning}.
At the same time, we notice two other methods~\cite{achlioptas2022changeIt3D, huang2022ladis} that leverage an additional edit instruction dataset for modifying shapes, but their codes are not publicly available.

To allow comparison, we prepare 50 exploration cases on 25 chairs and 25 tables in ShapeNet~\cite{chang2015shapenet}, each shape associated with a pair of description texts (input and target).
These inputs are given to our method and baselines for producing a modified shape.
Note that the input shape description is only used in our method and TextDeformer~\cite{Gao_2023_SIGGRAPH},
while the remaining methods can only take the target shape description.
Note that we also use a fixed step size $\alpha$ along the exploration direction. 

\paragraph{Sketch-guided.} 
We evaluate our framework against three recent sketch-based reconstruction baselines: Sketch2Model~\cite{zhang2021sketch2model}, Sketch2Mesh~\cite{guillard2021sketch2mesh}, and SketchSampler~\cite{gao2022sketchsampler}, which are designed for 3D reconstruction from sketch images.
We train one model on each category for each baseline using their officially-released codes.
As SketchSampler~\cite{gao2022sketchsampler} outputs only point clouds, we employ~\cite{Peng2021SAP} to produce meshes from their outputs.

To enable comparison, we rendered 50 sketch images (25 chairs and 25 tables) from the ShapeNet dataset~\cite{chang2015shapenet}.
As each baseline uses different rendering methods to prepare sketch images, we adopted their approaches to obtain their sketch images.
We first edited our sketch image in each exploration case, then modified the location in their sketch images to ensure consistency.
Some example sketch images taken by different methods are provided in supp.
In this exploration mode, we employ mechanism (i) in Section~\ref{ssec:dir_explore} to determine the step size $\alpha$.

\paragraph{User evaluation metrics.} 
We further invited 10 participants 
to assess the visual quality of the produced shapes and how well they match the input conditions; these two considerations are referred to as (i) quality score (QS) and (ii) matching score (MS).
During the evaluation, we showed to the participants a rendered image of each input shape and its associated exploration condition. 
We then presented the rendered images of the shapes produced by our method and by the baselines in a random order.
For each produced shape, we asked each participant to give ratings on QS and MS using a Likert scale from 1 (worst) to 5 (best).

\subsection{Quantitative Comparison}

\begin{table}[t]
\scriptsize
	\centering
		\caption{
Comparing our method with baselines in three exploration modes: (i) binary-attribute-guided, (ii) text-guided, and (iii) sketch-guided.
We can see that shapes generated by our method have the best quality for all metrics: lowest Frechet Inception Distance (FID), highest CLIP-Similarity (CLIP-S), highest Quality Score (QS), and highest Matching Score(MS).
 }
		\vspace*{-2mm}
		\resizebox{1.0\linewidth}{!}{
		\begin{tabular}{C{2.0cm}|@{\hspace*{0.3mm}}C{1.7cm}@{\hspace*{0.3mm}}|@{\hspace*{0.3mm}}C{0.8cm}@{\hspace*{0.6mm}}C{0.8cm}@{\hspace*{0.6mm}}C{0.5cm}@{\hspace*{0.6mm}}C{0.8cm}@{\hspace*{0.3mm}}}
			\toprule[1pt]
                Exploration scenario & Method & FID $\downarrow$ & CLIP-S $\uparrow$ & QS $\uparrow$ & MS $\uparrow$ \\ \hline
			\multirow{2}{*}{\shortstack[c]{Binary-attribute-guided}} & SVM in shape space
   & 156.0 & --- & 2.442  & 2.448 \\ 
			& Ours & \textbf{136.0} & --- & \textbf{4.176}  & \textbf{3.998} \\ \hline
            \multirow{4}{*}{\shortstack[c]{Text-guided}} & CLIP-Mesh
            & 255.5 & 0.676 & 1.140 & 1.192 \\ 
            & Text2Mesh
            & 214.2 & 0.768 & 2.068 & 1.596 \\
            & TextDeformer
            & 174.2 & 0.815 & 2.322 & 2.008  \\
            & Ours & \textbf{124.0} & \textbf{0.878} & \textbf{4.402} & \textbf{4.012} \\
            \hline
            \multirow{4}{*}{\shortstack[c]{Sketch-guided}} & Sketch2Model
            & 169.7 & 0.921 & 2.744 & 2.474 \\
            & Sketch2Mesh
            & 155.2 & 0.938 & 3.130 & 2.922 \\
            & SketchSampler
            & 198.5 & 0.909 & 2.060 & 2.092 \\
            & Ours & \textbf{124.4} & \textbf{0.976} & \textbf{4.680}& \textbf{4.620} \\ 
			\bottomrule[1pt]
	\end{tabular}}
	\label{tab:man_quanti}
	\vspace{-2mm}
\end{table} 
\paragraph{Evaluation metrics.}
Besides the above two user evaluation metrics, we compute 
the Frechet Inception Distance (FID), following~\cite{liu2022iss}, to evaluate the visual quality of the shapes produced by different methods.
Also, we measure the consistency of the produced shapes with the given conditions in the CLIP space, referred to as CLIP-Similarity (CLIP-S).
In the text-guided mode, we follow~\cite{fu2022shapecrafter} to compute this metric to measure the similarity between the produced shape's rendered image and the target shape description.
In the sketch-guided mode, we measure the score using the rendered sketch image of the produced shape and the given edited sketch image, following~\cite{zheng2023lasdiffusion}.
Since the binary-attribute-guided mode has no explicit condition,~\eg, sketch image or text, the CLIP-S metric is not applicable.
\vspace{2mm}

Table~\ref{tab:man_quanti} reports the quantitative comparison results on the three exploration modes.
In all three modes, our proposed framework can produce shapes with higher visual quality (confirmed by FID and QS), while being more consistent with the given conditions (measured by CLIP-S and MS).

\subsection{Qualitative Comparison}

\begin{figure*}[h]
	\centering
	\includegraphics[width=0.8\linewidth]{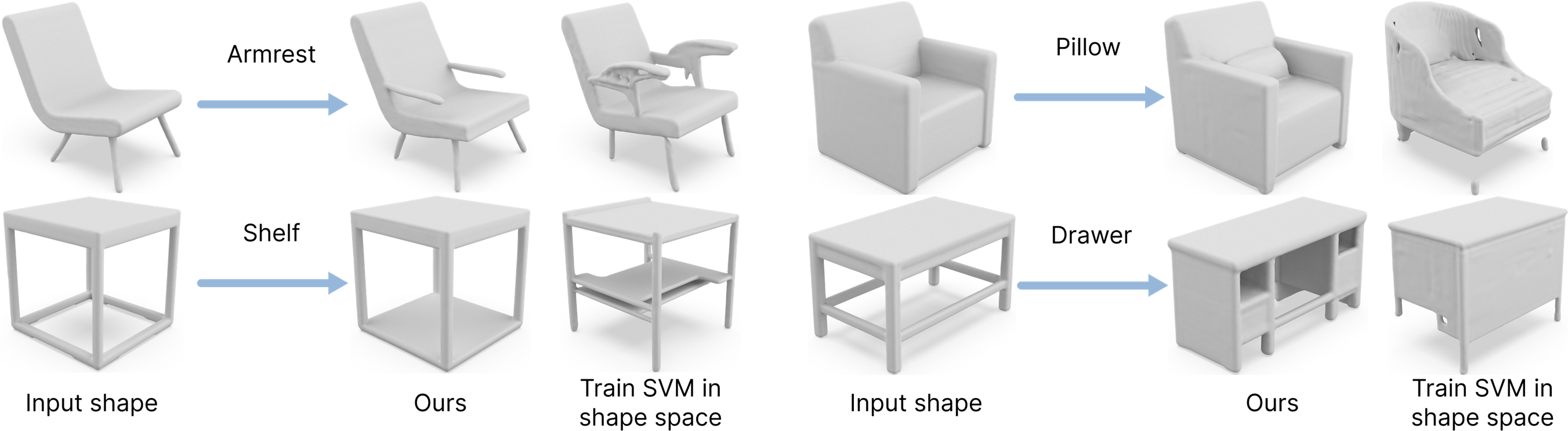}
	\vspace{-2mm}
	\caption{
            Visual comparison on the binary-attribute-guided exploration.
            We compare our framework with a modified version to directly train SVM in shape space instead of CLIP space (suggested in~\cite{zheng2022sdfstylegan}).
            Our framework can produce results that better follow the attributes (pillow and drawer) with fewer visual artifacts (armrest and shelf).
	}
	\label{fig:binary_comp}
	\vspace{-2mm}
\end{figure*}

\paragraph{Binary-attribute-guided.} 
Figure~\ref{fig:binary_comp} shows a visual comparison with against training SVM in shape space as described in~\cite{zheng2022sdfstylegan}.
Our proposed framework can produce more meaningful results from the given binary attribute than training SVM in the shape space; see the pillow and drawer on the right-hand side.
Furthermore, operating on the CLIP space following our framework can produce results with higher visual quality,  such as the armrest and the shelf on the left-hand side.

\begin{figure*}[h]
	\centering
	\includegraphics[width=0.99\linewidth]{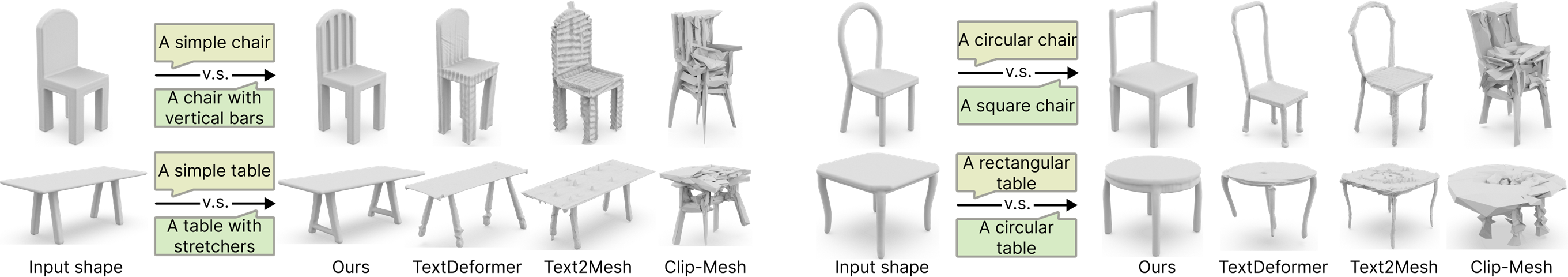}
	\vspace{-4mm}
	\caption{
            Visual comparison on the text-guided exploration.
            We compare our explored shapes with the one produced by TextDeformer~\cite{Gao_2023_SIGGRAPH}, Text2Mesh~\cite{oscartext2mesh}, and CLIP-Mesh~\cite{mohammad2022clip}.
            Our framework can change topologies based on a target shape description while existing deformation-based methods struggle with this.
            Our method can also produce more plausible shapes even when topology changes are not involved.
	}
	\label{fig:text_man_comp}
	\vspace{-2mm}
\end{figure*}

\paragraph{Text-guided.} 
As demonstrated in the left-hand side of Figure~\ref{fig:text_man_comp}, our text-guided shape exploration can modify the input shape's topology, such as vertical bars or stretchers, which pose challenges for other deformation-based methods.
For cases where the exploration does not introduce topology modification, as shown in the right-hand side of Figure~\ref{fig:text_man_comp}, our method can produce a shape that matches the target shape description.
Meanwhile, other baselines encounter difficulties in producing shapes that match the target shape description, or deformed shapes' visual quality is  limited.

\begin{figure*}[h]
	\centering
	\includegraphics[width=1.0\linewidth]{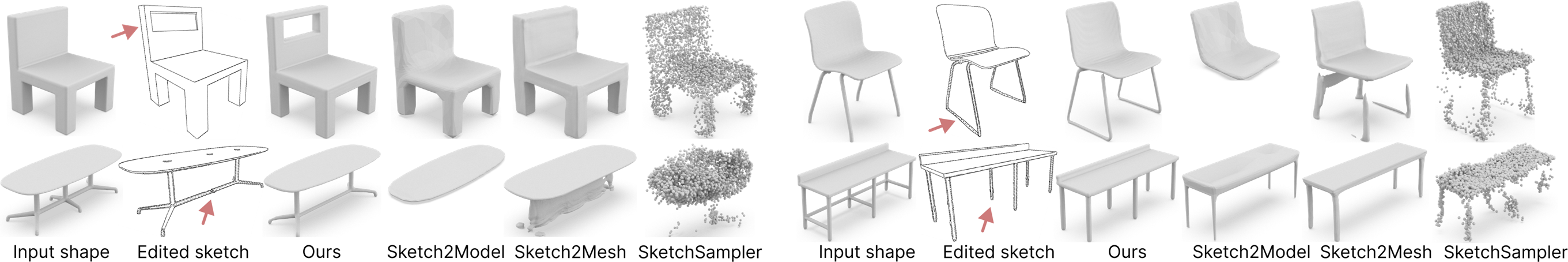}
	\vspace{-4mm}
	\caption{
            Visual comparison on the sketch-guided exploration, where changes in the sketch image are highlighted by red arrows.
            We compare our explored shapes with Sketch2Model~\cite{zhang2021sketch2model}, Sketch2Mesh~\cite{guillard2021sketch2mesh}, and SketchSampler~\cite{gao2022sketchsampler} show that our methods can better produce some thin structures (the table and chair on the right-hand side), while also being able to generate unusual structures, such as the hole on the chair back (upper-left) and the table's middle legs (bottom-right).
	}
	\label{fig:sketch_comp}
	\vspace{-2mm}
\end{figure*}

\paragraph{Sketch-guided.}
Figure~\ref{fig:sketch_comp} shows the visual comparison of our sketch-guided exploration with other baselines.
It is noted that our explored shapes can generate some thin structures on the shapes, such as the thin legs (the chair and table on the right-hand side), the other methods might fail to produce these structures.
Besides, we can notice that our method can produce some unusual structural patterns, \eg, the hole on the chair back, which are struggled by existing methods.

\vspace{1mm}
More comparisons are available in the supplementary material.

\subsection{More visual exploration results}
\begin{figure*}[h]
	\centering
	\includegraphics[width=1.0\linewidth]{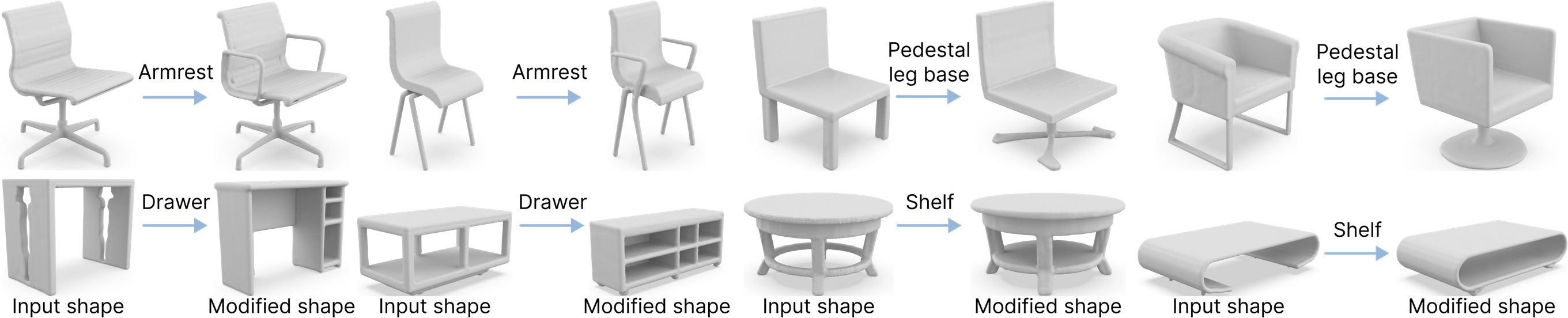}
	\vspace{-6mm}
	\caption{
            More results on binary-attribute-guided exploration. Our explored shapes are produced to match the binary attribute conditions; see the armrest growing from the chair and the drawer/shelf growing on the tables.
	}
	\label{fig:binary_man}
	\vspace{-2mm}
\end{figure*}

\begin{figure*}[h]
	\centering
	\includegraphics[width=1.0\linewidth]{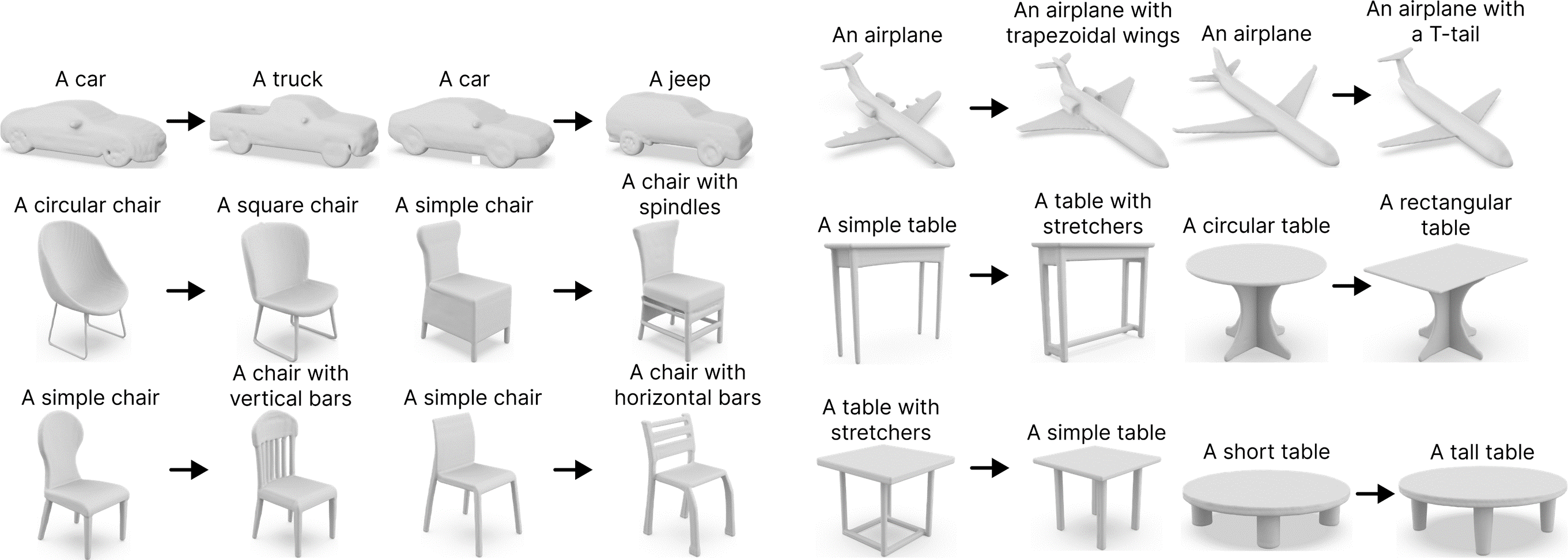}
	\vspace{-3mm}
	\caption{
            More results on text-guided exploration.
            By providing a diverse set of target shape descriptions, we can explore various input shapes to produce results that satisfy the corresponding description, \eg, the overall appearance of shapes (truck and jeep, short v.s. tall tables), and various topological structures (spindles/stretchers).
	}
	\label{fig:text_man}
	\vspace{-2mm}
\end{figure*}

\begin{figure*}[h]
	\centering
	\includegraphics[width=0.90\linewidth]{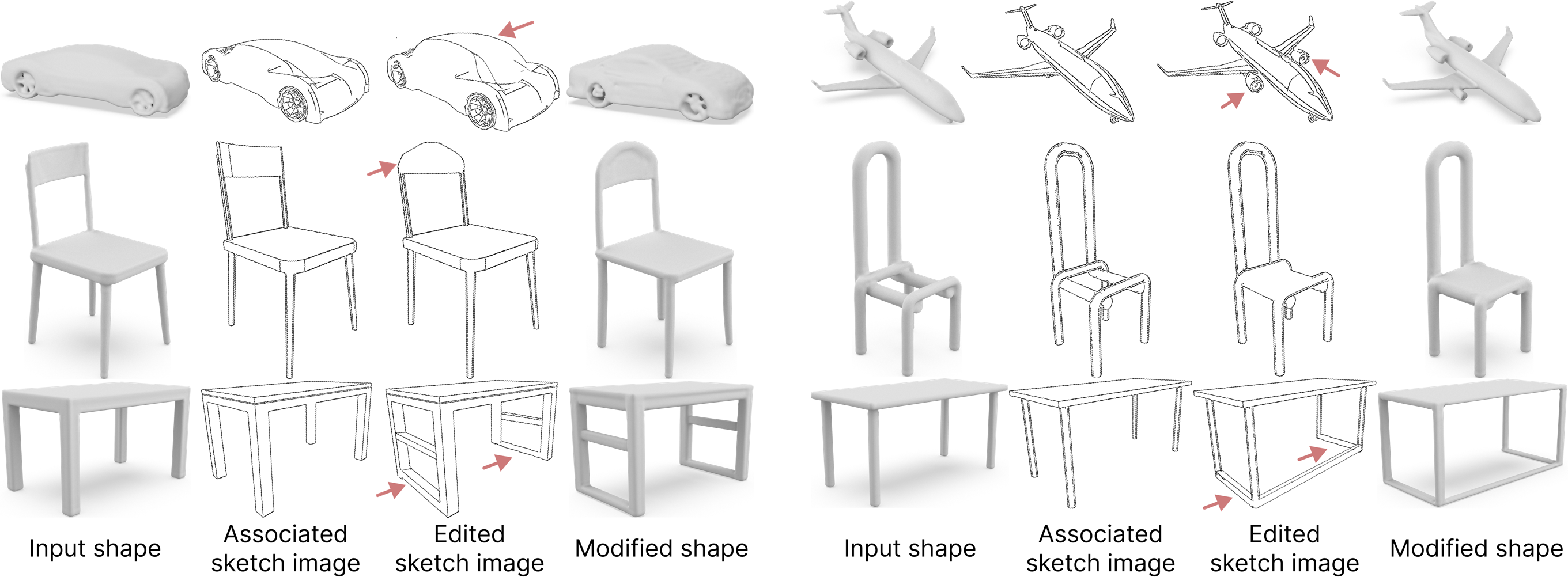}
	\vspace{-2mm}
	\caption{
            More results on sketch-guided exploration.
            Our framework can support changing the shape's local properties, such as adding engines to the airplane and changing the top of the car.
            We can also add or remove some local structures, such as stretchers on tables.
	}
	\label{fig:sketch_man}
	\vspace{-2mm}
\end{figure*}

Figures~\ref{fig:binary_man},~\ref{fig:text_man}, and~\ref{fig:sketch_man} demonstrate the capacity of our framework to explore meaningful modifications of input shapes based on user-specified conditions.
Specifically, we present visual results from three supported exploration modes.
Overall, our proposed framework can take multimodal conditions to produce meaningful exploration directions.
The direction enables the user to explore the input shape and produce high-fidelity shapes that satisfy the associated condition.
More exploration results are provided in supp.

\subsection{Model Analysis}
Please refer to the supplementary material for the effectiveness analysis of our CLIP2Shape mapper via sketch reconstruction task and the ablation study of our co-optimization strategy.

\section{Conclusion}
In this work, we propose the first framework for exploring a pre-trained 3D shape space using the vision-language model, specifically the CLIP model.
Our approach connects the CLIP space with the pre-trained space via an additional trained CLIP2Shape mapper.
We then identify a suitable CLIP code for representing the input shape and locate a linear exploration direction in CLIP space.
Our framework supports locating relevant exploration trajectories according to multi-modal conditions, such as binary attributes, texts, and sketch images. 
Our experimental results demonstrate our method is able to produce various meaningful explored shapes, which are challenging for existing methods.
Furthermore, we have confirmed the coherent connection of the latent spaces and the effectiveness of the co-optimization strategy via various model analyses.

\paragraph{Limitation and discussion.}

First, following the exploration trajectory may lead to unexpected changes in the shape beyond the given conditions.
Second, although an exploration trajectory is used, tracing with a suitable step size is still not trivial.
Last, since the two pre-trained spaces are generally black boxes, it can be difficult to determine whether an input condition falls within the capabilities of the pre-trained models.
Please refer to the supplementary material for a more detailed discussion on the limitation with examples.

\if 0
\fi

\bibliographystyle{ACM-Reference-Format}
\bibliography{bibliography}

\clearpage

\end{document}